%% file: main.tex
\documentclass[10pt,twocolumn,letterpaper]{article}

\input{preamble}

\usepackage[a4paper, total={6.5in, 8.5in}]{geometry}

\usepackage[breaklinks,colorlinks]{hyperref}

\title{Kandinsky Conformal Prediction: Efficient Calibration of Image Segmentation Algorithms}

\author{ Joren Brunekreef$^{* \, 1,2}$ 
	     \ Eric Marcus$^{* \, 1,2}$ 
        \ Ray Sheombarsing$^{ 1}$ 
	     \ Jan-Jakob Sonke$^{1,2}$ 
	     \ Jonas Teuwen$^{1,2}$ 
	     \\[2ex]      
	      \emph{\normalsize{$^1$The Netherlands Cancer Institute}} \\
       	       \emph{\normalsize{Amsterdam, Plesmanlaan $121$, $1066$ CX, The Netherlands}} \\[2ex]
       		\emph{\normalsize{$^2$University of Amsterdam}} \\
       		\emph{\normalsize{Amsterdam, Science Park $900$, $1012$ WX, The Netherlands}	  }   
}
\begin{document}
\maketitle
\def\thefootnote{*}\footnotetext{These authors contributed equally to this work. Corresponding emails: j.brunekreef@nki.nl, e.marcus@nki.nl.}
\begin{abstract}
\emph{Image segmentation algorithms can be understood as a collection of pixel classifiers, for which the outcomes of nearby pixels are correlated. Classifier models can be calibrated using Inductive Conformal Prediction, but this requires holding back a sufficiently large calibration dataset for computing the distribution of non-conformity scores of the model's predictions. If one only requires only marginal calibration on the image level, this calibration set consists of all individual pixels in the images available for calibration. However, if the goal is to attain proper calibration for each individual pixel classifier, the calibration set consists of individual images. In a scenario where data are scarce (such as the medical domain), it may not always be possible to set aside sufficiently many images for this pixel-level calibration.
The method we propose, dubbed ``Kandinsky calibration'', makes use of the spatial structure present in the distribution of natural images to simultaneously calibrate the classifiers of ``similar'' pixels. This can be seen as an intermediate approach between marginal (imagewise) and conditional (pixelwise) calibration, where non-conformity scores are aggregated over similar image regions, thereby making more efficient use of the images available for calibration. We run experiments on segmentation algorithms trained and calibrated on subsets of the public MS-COCO and Medical Decathlon datasets, demonstrating that Kandinsky calibration method can significantly improve the coverage. When compared to both pixelwise and imagewise calibration on little data, the Kandinsky method achieves much lower coverage errors, indicating the data efficiency of the Kandinsky calibration.}
\end{abstract}

\section{Introduction}
Calibration of predictive models is a critical aspect of machine learning, particularly in applications with significant impact based on model outcomes, such as medical diagnostics. Calibration ensures that predicted probabilities match the actual empirical likelihood of the predicted events. A well-calibrated model will output probabilities that correspond closely to real-world frequencies; for instance, if a model predicts an event with a probability $p$, this event should, in reality, occur with frequency $p$.

In this work, we will focus on image segmentation tasks. Here, the calibration procedure also plays a vital role. A segmentation model can be interpreted as a collection of classifiers, one for each output pixel. The calibration of each classifier thus affects decision-making at the pixel level, which subsequently influences global measures of segmentation accuracy (e.g. the Dice score). There are two straightforward notions of calibration in the context of image segmentation: \emph{marginal} calibration, which measures the calibration averaged over all pixels, or calibration \emph{conditional} on a specific pixel location. 

One standard method for calibrating prediction models is Conformal Prediction (CP), a framework that has received increasing amounts of attention in recent years \cite{angelopoulos2022conformal,angelopoulos2022learn,angelopoulos2023predictionpowered,kutiel2023conformal,romano2020classification,angelopoulos2022imagetoimage}. Conformal prediction provides statistically valid measures of confidence in a model-agnostic manner. Whereas the original (transductive) CP method \cite{saunders1999transduction} is computationally demanding, the more recently developed \emph{inductive} CP \cite{papadopoulos2002inductive,papadopoulos2007conformal,papadopoulos2008inductive,vovk2012conditional} is better suited for a present-day machine-learning setting. This lower demand for computational resources comes at the cost of requiring extra (labeled) data to be set aside as a calibration set. One then defines a notion of a \emph{non-conformity score}, which measures the ``strangeness'' of each sample and the associated model prediction in this calibration set. The distribution of non-conformity scores in the calibration set can then be used as a benchmark to compare with newly unseen samples, which allows one to obtain a statistically valid notion of confidence for the model's predictions.

Now consider the case where one has set aside a calibration set of $N$ images of dimensions $m \times n$. If we require only marginal calibration (i.e. aggregated over the whole image), we can view each pixel in each image as a separate calibration data point, so that our calibration set is of size $N \times m \times n$. However, since a segmentation model contains a ``separate'' classifier for each individual pixel, calibrating such a model on the pixel level requires one to view each image in the calibration set as a single sample. The calibration set is, therefore, only of size $N$ in this case. Clearly, this can pose a challenge for settings where data availability is limited --- for example, in the medical domain.

To address these challenges, we introduce ``Kandinsky calibration'', a technique that capitalizes on prior knowledge of the spatial correlations within images to calibrate classifiers across similar pixels more efficiently. This approach balances the need for detailed calibration at the pixel level with the practical limitations of data availability. It applies conformal prediction in a novel way, achieving fine-grained calibration with fewer calibration images. The following sections will detail the Kandinsky calibration approach and present experiments demonstrating its effectiveness in improving the calibration when few calibration images are available, a valuable attribute for segmentation applications.

\paragraph{Related Work} Calibration for machine learning, and in particular deep learning, has received much attention of late \cite{guo2017calibration,minderer2021revisiting,carrell2022calibration,conde2023theoretical,kuppers2022confidence}. For example, the ``formalization'' of the calibration has been discussed in \cite{blasiok2022unifying}, providing the first steps to a more formal understanding of the procedure and of calibrated functions. Conformal prediction \cite{vovk2005algorithmic,papadopoulos2002inductive,lei2014distribution, angelopoulos2021gentle}, the framework we utilize in this work, is also growing increasingly popular. Although work on risk control for segmentation purposes has been studied \cite{angelopoulos2022conformal}, thorough investigations of calibration methods for segmentation are scarce \cite{wieslander2021deep}.

\paragraph{Overview} The contributions of this work are ordered as follows. In \autoref{sec:calibration_conformal}, we provide a short overview of calibration, conformal prediction, and coverage. The following section, \autoref{sec:kandinsky_calibration}, we discuss our novel Kandinsky calibration framework. In \autoref{sec:experiments}, we show the experimental results of the different calibration methods.

\section{Calibration and Conformal Prediction}\label{sec:calibration_conformal}
Due to the significant role of conformal prediction, calibration, and coverage, we provide a short introduction containing relevant information for this work.

\subsection{Conformal Prediction} We provide a short introduction of conformal prediction \cite{vovk2005algorithmic,papadopoulos2002inductive,lei2014distribution}, following the conventions of the excellent introduction in \cite{angelopoulos2021gentle}.

Let us consider a classification task first. Suppose we are given a training set of images and labels of $K$ classes. Furthermore, we train a predictive model $ f $ on this data, such that its outputs $ f(x) \in [0,1]^K$. Using the inductive conformal prediction framework, we then apply the model $ f $ to so-called calibration data, consisting of $n$ i.i.d. unseen samples $I = (X_1, Y_1), \ldots, (X_n, Y_n)$. 

Utilizing this calibration data and $f$, we consider a new (unseen) datapoint $(X_{\text{test}}, Y_\text{test})$, where we do not know $Y_\text{test}$. The objective is to create a prediction set $\mathcal{C}_\alpha(X_{\text{test}}) \subset \{1,2,\dots,K \}$ with the following property
\begin{equation}
P(Y_{\text{test}} \in \mathcal{C}_\alpha(X_{\text{test}})) \geq 1-\alpha \ ,
\label{eq:conf-guarantee}
\end{equation}
where $\alpha$ is a user-chosen error rate. We can create these prediction sets by defining a so-called non-conformity score $s(x, y)$ that measures how far off the model's prediction $y$ on an input $x$ is from the ground truth. Then we define $\hat{q}_\alpha$ as the $\lceil (n+1)(1-\alpha) \rceil / n$ empirical quantile of the $s_i \equiv s(X_i, Y_i)$ in the calibration set. For the new test point $X_\text{test}$ (where the label is unknown), we create the prediction set
\begin{equation}
    \mathcal{C}_\alpha(X_{\text{test}}) = \{ y \, | \, s_i \leq \hat{q}_\alpha \} \ .
\end{equation}
It can then be shown that $\mathcal{C}_\alpha(X_{\text{test}})$ satisfies \eqref{eq:conf-guarantee}.

The choice of the scoring function $s(x, y)$ determines the usefulness of the prediction sets, and what choice to make here depends on the task at hand. For the remainder of this work, we set
\begin{equation}
    s(x, y) := 1 - f(x)_Y,
\end{equation}
where the subscript $Y$ indicates that we take the model's output for the ground truth class $Y$. We leave it for future work to investigate whether other scoring functions could lead to better performance in the context of segmentation.

\paragraph{Segmentation} In the case of segmentation, the model can be seen as a collection of classifiers, one for each pixel. Since these are separate classifier models, they should in principle be calibrated independently if we want to create valid prediction sets for each individual pixel. The upshot of this is that every labeled image $(X_i, Y_i)$ in a calibration dataset $I$ should be taken as just a single calibration sample for which we compute the non-conformity score $s_i$. We call this \emph{pixelwise} (or \emph{conditional}) calibration. 

If, however, we set ourselves the more modest goal of \emph{marginal} calibration where we only need the prediction sets to be valid on average over the whole image, we can view each pair $(X_i, Y_{i,(x,y)})$ of input images with the ground truth value for the pixel with coordinate $(x,y)$ as a separate calibration point. However, marginal calibration can be attained even if a number of individual pixels are severely miscalibrated, so this approach is suboptimal if we want to have proper calibration guarantees in specific regions of the image. 

If, therefore, one has access to a sufficiently large set $I$ of calibration images, it is preferable to apply the pixelwise calibration method. However, since labeling data for segmentation is a time-consuming task, such data are often scarce.
\subsection{Calibration and Coverage Errors}

\begin{figure}
    \centering
    \includegraphics[width=0.8\columnwidth]{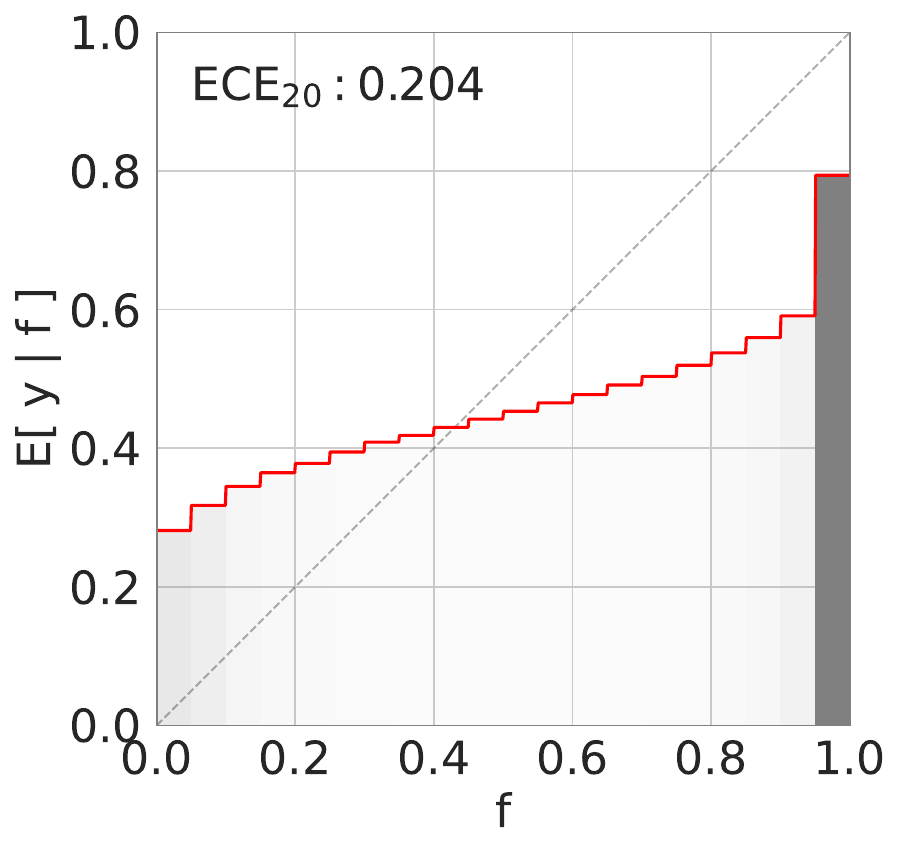}
    \caption{Reliability diagram \cite{blasiok2023smooth} of a segmentation model trained on a subset of MS-COCO. The model's prediction scores and associated accuracies are aggregated over all pixels and assigned to 20 bins of equal width. The ECE is computed by averaging the absolute difference between the height of the bins and the diagonal. This particular model is overconfident for output scores $f \gtrapprox 0.4$ and underconfident for lower output scores.}
    \label{fig:reliability-diagram}
\end{figure}

A common calibration measure for prediction models is the Expected Calibration Error (ECE). For a binary classification task, it can be defined as 
\begin{equation}
    \textrm{ECE}_{\mathcal{D}}(f) = \mathbb{E}_\mathcal{D} \left[ \left| \mathbb{E}_{y \mid f(x)} [ f(x) ] - f(x) \right| \right].
\end{equation}
Here $\mathcal{D}$ is the data distribution on inputs $x$ and ground truth labels $y \in \{0, 1\}$. This expression, however, is not well-defined for finite samples since it conditions on null events. The ECE can be approximated in several ways \cite{blasiok2022unifying}. One particularly convenient and efficient method is to collect the model output scores $f(x_i)$ (evaluated on a labeled dataset of $N$ samples $x_i$) in bins $B_m$, and compute the so-called binned ECE:
\begin{equation}
    \textrm{bECE}_M= \frac{1}{N} \sum_{m=1}^M \left|B_m\right| \left(\textrm{conf}\left(B_m\right) - \textrm{acc}\left(B_m\right)\right),
\end{equation}
where $\textrm{conf}\left(B_m\right)$ is the mean output score in the bin $B_m$ and $\textrm{acc}\left(B_m\right)$ is the true fraction of positive samples for which the score $f(x_i)$ is assigned to this bin. Throughout this work, we refer to the binned ECE simply as $\textrm{ECE}_M$ (with a subscript indicating the number of bins) to avoid clutter. We present an example of a so-called \emph{reliability diagram} in \ref{fig:reliability-diagram}, where we plot the accuracy as a function of the confidence of the prediction model used in one of our experiments.

The ECE for a perfectly calibrated model equals zero: the model's output scores are precisely equal to the corresponding observed accuracies in the long run. Note that this does not imply the model has good classification performance: a  prediction model that outputs a score of $1/2$ for each input is perfectly calibrated if exactly half the samples belong to the positive class.

When calibrating a model using conformal prediction, we need another measure of calibration performance. The aim of the prediction sets $\mathcal{C}_\alpha$ is to satisfy \emph{coverage} conditions, meaning that the correct (ground truth) class $Y_\text{test}$ of an unseen input $X_\text{test}$ should be in the prediction set $\mathcal{C}_\alpha\left(X_\text{test}\right)$ with probability at least $1 - \alpha$. We therefore define a (binned) measure of the \emph{Coverage Error} (CE) as follows:
\begin{equation}
    \textrm{CE}_M = \sum_{m=1}^M \left(\textrm{cov}\left(\mathcal{C}_{1 - \frac{m}{M}}\right) - \frac{m}{M} \right).
\end{equation}
The coverage $\textrm{cov}\left(\mathcal{C}_\alpha\right)$ is defined as the frequency with which the $\alpha$ level prediction set $\mathcal{C}_\alpha$ contains the ground truth class. We point out to the reader to carefully distinguish between the abbreviations `ECE' for Expected Calibration Error and `CE' for Coverage Error. 

\begin{figure}
    \centering
    \includegraphics[width=0.8\columnwidth]{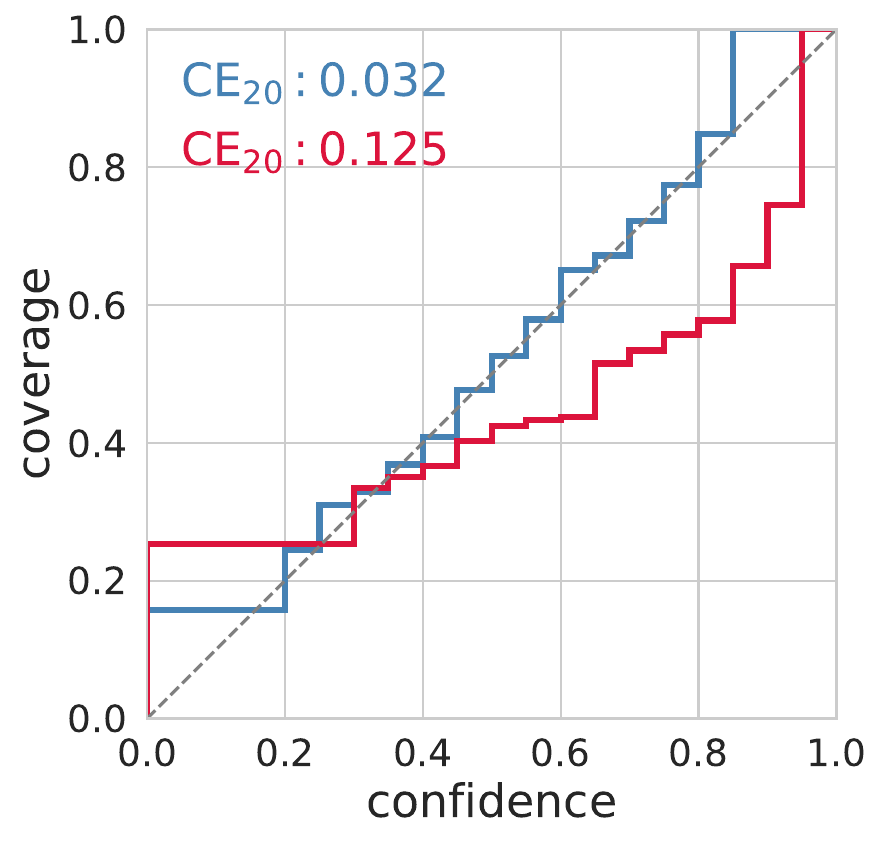}
    \caption{Coverage diagram for two individual pixel locations of a segmentation model trained on a subset of MS-COCO, calibrated pixelwise on 20.000 images (blue) and on 100 images (red).}
    \label{fig:coverage-diagram}
\end{figure}

For all our experiments in this work, we report the $\textrm{CE}_{20}$, that is, the coverage error obtained with 20 bins.
A \emph{coverage diagram} is the counterpart of a reliability diagram when evaluating the coverage properties of our prediction sets. In \ref{fig:coverage-diagram} we present the ``coverage curves'' of two of the (partially) calibrated pixel classifiers used in one of our experiments.

\section{Kandinsky Calibration}\label{sec:kandinsky_calibration}
Now that we have discussed the two `naive' approaches of marginal and conditional calibration, we introduce our Kandinsky method. The primary rationale is as follows: \emph{the pixel-level classifiers of a segmentation network are not independent and are related by well-organized patterns encoded in the task and data}. In other words, we are able to use prior knowledge in the calibration process of the segmentation networks.

In order to do this, we first define the notion of a non-conformity \emph{curve}. Once we have computed non-conformity scores for a classifier model on a calibration set, we use its $\hat{q}$-th quantiles to create prediction sets for unseen data. This $\hat{q}$-th quantile of the non-conformity scores is what we refer to as the non-conformity curve $z(\hat{q})$, where $0 \leq \hat{q} \leq 1$. Note that non-conformity curves are defined for all $\hat{q}$ in this range, even if it based on only a small number of individual non-conformity scores.


When performing pixelwise calibration, we compute a separate non-conformity curve for each pixel coordinate in the image. The other extreme, imagewise calibration, computes a single non-conformity curve (based on the non-conformity scores aggregated over the whole image) that is then used for all pixel locations. Our Kandinsky method is an intermediate approach, where we cluster nearby pixels and compute a non-conformity curve for each cluster by aggregating their non-conformity scores. 

This clustering is performed by finding pixels with similar non-conformity curves. This approach may seem circular, since we first require the (potentially noisy) non-conformity curves themselves to subsequently improve these curves over the whole cluster. However, the prior knowledge used in forming our clusters is that spatially nearby pixels are likely to have similar non-conformity curves: therefore, even if individual pixel locations in a certain spatial region have a dissimilar non-conformity curve due to a lack of data, they will still be grouped together with other pixels in their neighborhood.

The general Kandinsky method for calibrating a prediction model $f$ can then be outlined as follows:
\begin{itemize}
    \item Perform pixelwise calibration, computing separate non-conformity curves for each pixel location. These non-conformity curves are likely to be noisy in a low-data scenario.
	\item Cluster pixel locations based on similarity of their non-conformity curves, potentially with a prior choice of possible region shapes.
	\item For each cluster, aggregate all the non-conformity scores encountered in the cluster, and compute a cluster-specific non-conformity curve.
	\item When forming prediction sets for a given pixel, use the newly obtained non-conformity curve of the cluster to which this pixel belongs.
\end{itemize}

\subsection{Computing Kandinsky Clusters}
Having motivated the creation of clusters of non-conformity curves, we will now provide several example methods of computing them.

\subsubsection{K-Means Clustering}
As a first approach, we can find Kandinsky clusters using a k-means clustering approach. In particular, we start with a pixel-level calibrated model and subsequently consider the non-conformity curves per pixel. At this stage, we choose a set of $k$ quantiles at which we wish to compare non-conformity curves between the different pixels. The k-means clustering approach then tries to group all $m \times n$ (image size) points in this $k$-dimensional space. Provided we can obtain `good' enough calibration measurements per pixel, this approach will yield precise information about which pixels are related.

In the low-data regime, however, this approach will be prone to noise in the calibration of the individual pixels. The resulting clusters will most likely be only based on spurious relations and provide no help in obtaining better calibration. To optimize groups even in this low-data regime, we propose two more methods.

\subsubsection{Genetic Algorithms}
Genetic algorithms \cite{holland1992adaptation,goldberg1989cenetic} are a class of optimization techniques that ``simulate'' the process of natural evolution. These algorithms excel in navigating complex search spaces to identify solutions that might otherwise be inaccessible through traditional optimization methods. The specific type of algorithm we will employ is the so-called differential evolution \cite{storn1997differential}. In the rest of the work, we refer to the approach we outline here as GenAnn (for Genetic Annuli).

To be precise, for GenAnn, we will need to define a fitness function $F$ that evaluates how `good' a candidate solution is, a crossover function $C$ which takes a collection of candidate solutions and combines them into a new one, a mutating function $M$ which takes a candidate solution and randomly transforms it, and a replacement function $R$ that determines if the candidate solutions are replaced with the mutated ones. Furthermore, we must provide a parametrization $x_i$ of a candidate solution in terms of a finite set of real parameters. In our experiments, this parametrization will consist of the center coordinates and a set of radii for annuli. We show the specific implementations of the above functions and general procedure used in differential evolution for the reader's convenience in the supplementary material, \autoref{algorithm:genetic_algo}.

\subsubsection{Fourier Concentric Clustering}
 Here, we introduce the \emph{Fourier Concentric Clustering} (FCC) method. In this section, we shall provide only a short summary of the method; a detailed and formal explanation can be found in the supplementary material, \autoref{appendix:fcc}. Formally, we wish to construct a concentric decomposition of $\mathcal{R}_{n} := [0, n_{1} - 1] \times [0, n_{2} -1]$ on which 
the variances of a user-prescribed quantity $\J: \mathcal{R}_{n} \rightarrow \RR^{p}$ are minimized. This approach systematically decomposes an image into a series of concentric, nested subsets \( V_{0} \subset \ldots \subset V_{m-1} \subset \mathcal{R}_{n} \), centralizing around the mean of $\J$ with minimized variance
\begin{equation}
 \sum_{l=0}^{m} \mathbb{E}_{A_{l}} \left( \Vert \mathcal{J} - \bm{\mu}_{l} \Vert^{2} \right) \ ,
\end{equation}
where $\bm{\mu}_{l}$ denotes the mean of $\J$ on $A_l$, and each $A_l$ will be analogous to the annuli discussed in the genetic algorithms. In other words, these $A_l$ will be one of the nested subsets $V_n$, excluding the previous $V_{n-1}$.

To facilitate the decomposition, we redefine the image domain \( \mathcal{R}_{n} \) in a new coordinate system, with the image's midpoint serving as the origin. The boundary of each subset \( A_{l} \) is then represented by polar curves, which are expressed via a finite Fourier series to allow for computational tractability. For example, the boundary of the initial subset \( A_{0} \) is parameterized as 
\begin{equation}
    A_{0} = \left\{ (r, \theta) \mid 0 \leq r \leq r_{0}(\theta), \theta \in [0,2\pi] \right\} \ ,
\end{equation}
with \( r_{0} \) being a smooth, periodic function.

The numerical integration necessary for evaluating the variances is achieved using Legendre and Fourier quadrature methods, providing a means to compute the integrals over the domains \( A_{l} \) as functions of the Fourier coefficients. This is exemplified by the equation \( \int_{A_{0}} f(x) \,dx = 2\pi c_{00} \), where \( c_{00} \) is the zeroth Fourier coefficient of a line integral, approximated using the FFT.

In setting up the optimization problem to minimize variance, we initialize the Fourier coefficients with a perturbation \( \varepsilon \) to represent a `noisy' set of concentric circles. The BFGS algorithm \cite{broyden1970convergence,shanno1970conditioning,fletcher1970new,goldfarb1970family} is then employed to find the optimal coefficients that minimize the variance term coupled with a regularization term \( \Lambda \) to enforce the nested nature of the subsets without overlapping. The FCC method allows for a faster optimization of circular and elliptical cases than the genetic approach. However, the genetic approach provides a much more flexible
       optimization process as opposed to the Fourier approach, where special care has to be taken not to ignore higher-order modes

\subsection{Limitations} The most prominent limitation lies in creating prior knowledge of the calibration. There is no universal way to make the Kandinsky clusters, as it depends deeply on the characteristics of the data and task; it will require human input. The concentric circles and ellipses will not work for every imaginable task. Secondly, the numerical methods we introduced to compute the Kandinsky clusters may not be able to handle all possible geometric priors. The genetic algorithm should, in principle, be able to optimize the result, provided a suitable parametrization of the objective can be found; this is, however, not always a trivial task.

\section{Experiments}\label{sec:experiments}
\begin{figure}
	\centering
	{{\includegraphics[width=0.8\columnwidth]{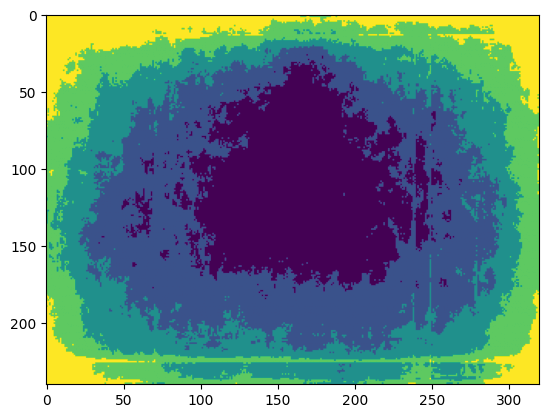}}} \\[1ex]
	\caption{Results of the k-means clusters of non-conformity curves. A U-Net model was calibrated on 20.000 calibration images. During this calibration procedure, we obtained non-conformity curves for each pixel output of the model for the `person' class. We obtained the non-conformity score for each curve for the 60, 70, 80, and 90th quantiles. These four-dimensional points were subsequently clustered using k-means. The appearance of concentric shapes is visible in the image. These shapes arise due to the different prevalences of persons appearing in different locations of the image. Furthermore, persons near the image's border are likely more difficult to identify because they might, for example, be found in the distance more often. The combination of the object (persons in this case) and the data characteristics determine the precise geometry of the clusters.
		     \label{fig:kandinsky_cluster_kmeans}
		   }
\end{figure}

\begin{figure}
	\centering
	{\includegraphics[width=0.8\columnwidth]{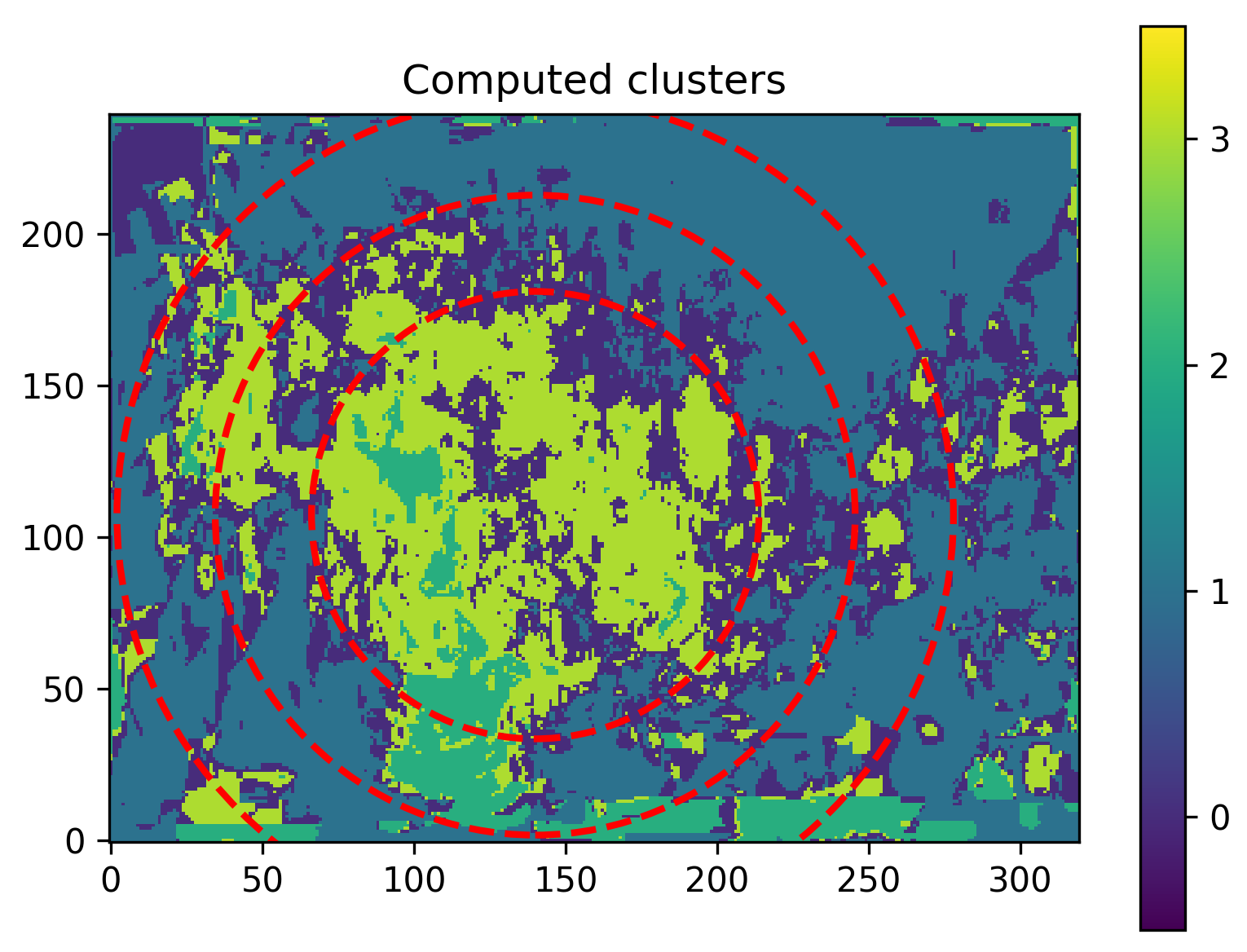}} \\[1ex]
	\caption{This figure shows the most `extreme' scenario, where we have only used 100 images from the MS-COCO dataset during the calibration. The k-means clusters are shown in the background, clearly unable to find the same clusters as in \autoref{fig:kandinsky_cluster_kmeans}. The red rings show the radii of the annuli found to minimize the objective using the genetic algorithm approach.
		     \label{fig:cluster_plot_100}
		   }
\end{figure}

\begin{figure}
    \centering
    \includegraphics[width=0.7\columnwidth]{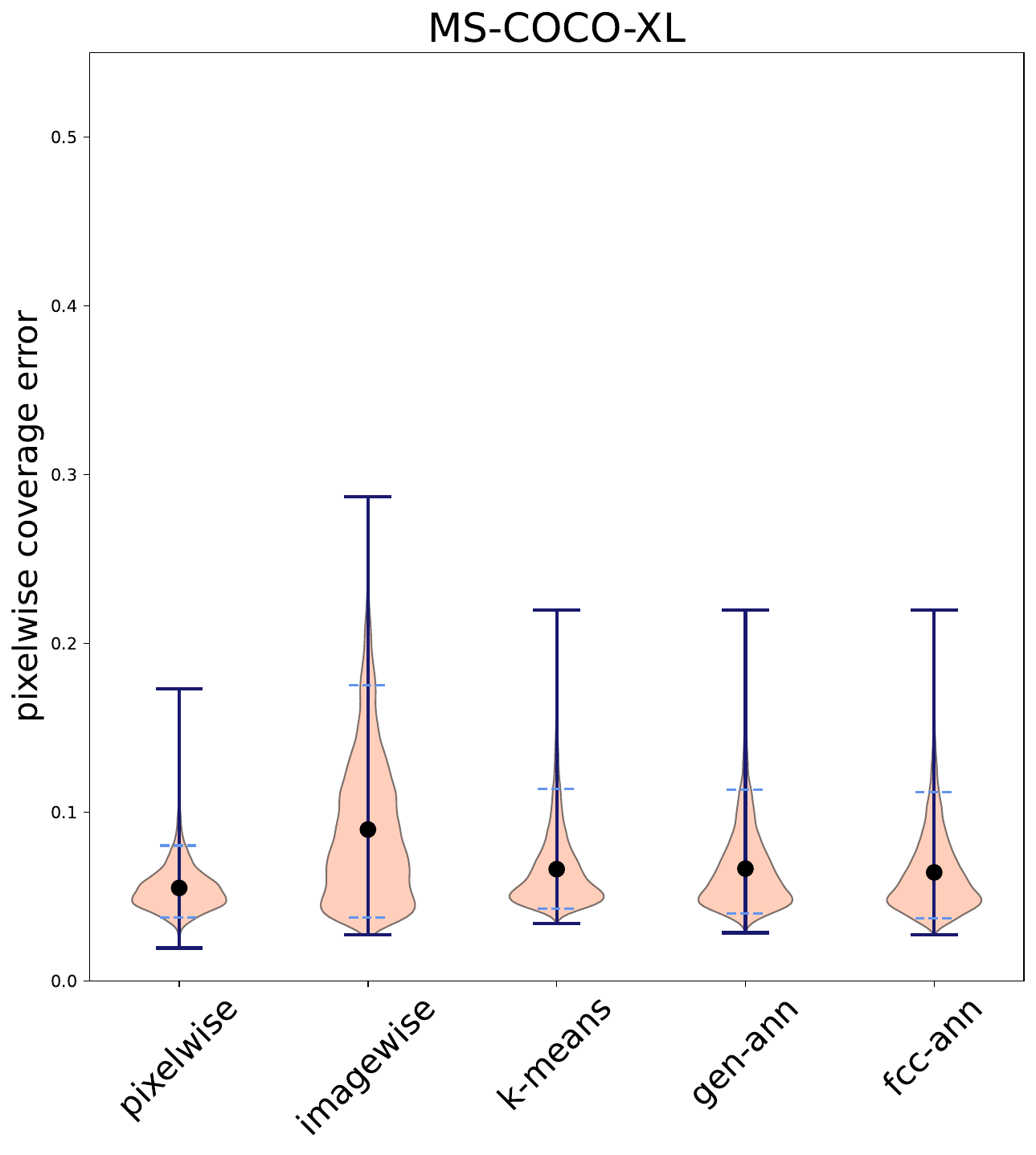}
    \caption{Violin plot of the pixelwise coverage errors for the MS-COCO-XL experiment. This experiment investigated the idealized scenario with access to large calibration sets. Here, we have utilized 20.000 images to calibrate the segmentation model. Due to the size of this calibration set, we can effectively use the pixelwise calibration, which attains the lowest mean coverage errors. The Kandinsky methods follow closely, and only the imagewise calibration performs visibly worse. The reason for this is that imagewise calibration will `average' over all the pixels, providing for each individual pixel a skewed estimate of its calibration.}
    \label{fig:violin_mscoco_xl}
\end{figure}

\begin{figure}
    \centering
    \includegraphics[width=0.7\columnwidth]{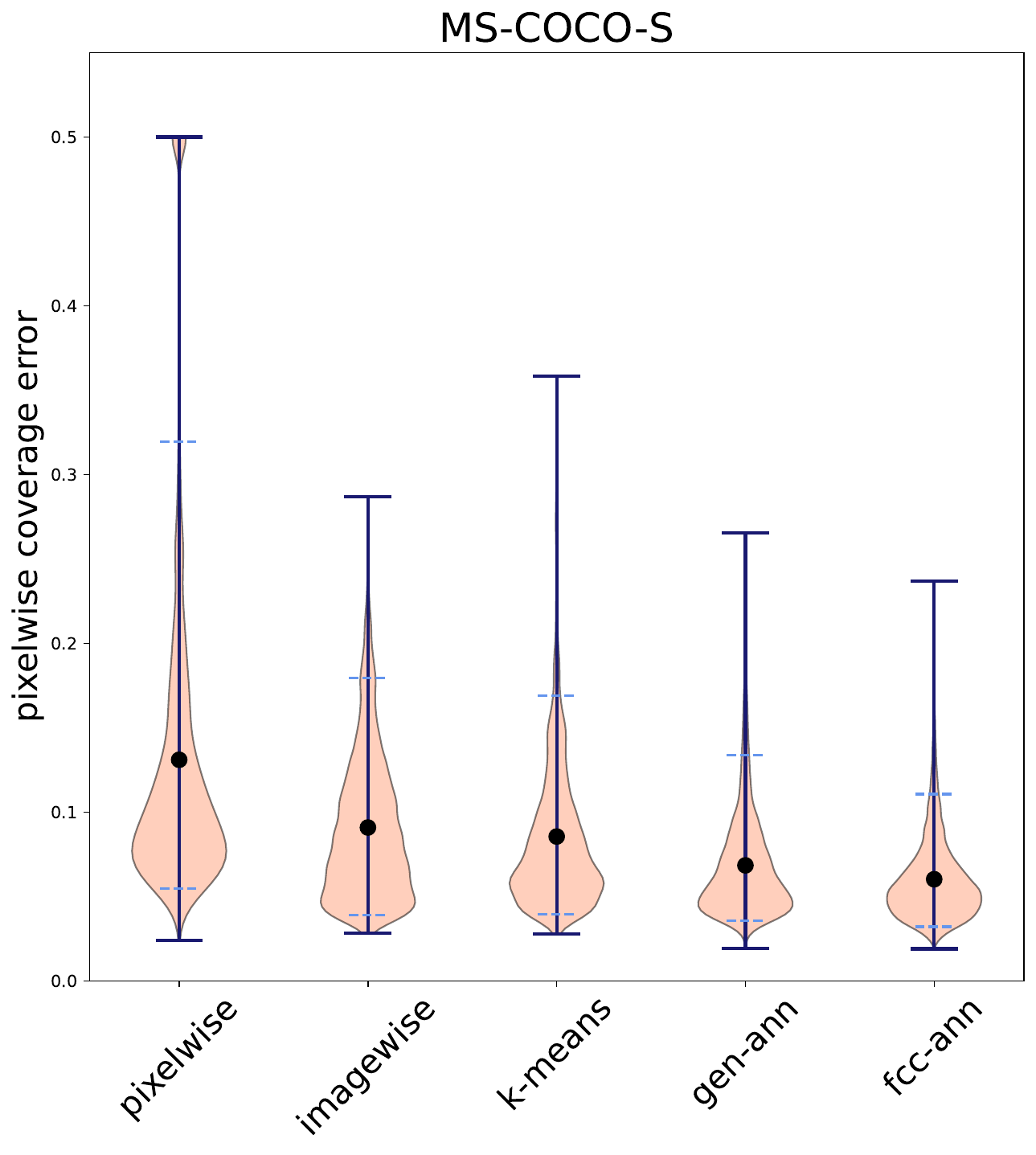}
    \caption{Results of all calibration methods using only 100 calibration images. All Kandinsky methods (the three rightmost) outperform the pixelwise and imagewise calibrations. The GenAnn and FCC annuli outperform the K-means approach due to the increased noise in the clusters found by K-means.}
    \label{fig:violin_mscoco_s}
\end{figure}

This section will investigate the utility of using Kandinsky clusters in the calibration procedure. We will compare the marginal, conditional, and Kandinsky methods for various trained models on the unseen test sets. In particular, we wish to investigate the coverage errors attained by the different models and methods, as that will provide us with a robust metric. 
As we have described, the Kandinsky methods will be most helpful in scenarios with little calibration data. To investigate this, we have set up four experiments with different data availabilities:

\begin{itemize}
    \item \textbf{MS-COCO-XL}: the highly idealized setting where we utilize 20.000 images for calibration.
    \item \textbf{MS-COCO-S}: same dataset, but we use only 100 calibration images.
    \item \textbf{Decathlon-L}: a, for medical standards, large calibration dataset consisting of 77 patients.
    \item \textbf{Decathlon-S}: same dataset, but utilizing only 27 patients for calibration.
\end{itemize}

The different objectives and calibration dataset sizes allow us to investigate the utility of Kandinsky calibration across varying scales and domains, providing insights into its effectiveness in more abundant and scarce data scenarios and its adaptability to diverse image contexts. In \autoref{tab:mean-ces}, we summarize the results of all the experiments, showing the mean coverage errors attained by all methods.

\paragraph{Model} For all experiments in this section, we utilize a U-Net for the segmentation. In particular, we have a U-Net with four up and downsampling layers, resulting in $53.5$M parameters. Since our goal lies with the calibration of the models and not finding the maximal possible performance of the models, there is no need for extensive hyperparameter searches. In the case of MS-COCO we utilized 678 datapoints for training and 2869 unseen images to evaluate the coverage. For Medical Decathlon, the model was trained on 86 patients and evaluated on 118 patients.

\subsection{MS-COCO-XL}
The goal of this first experiment is to investigate a highly idealized scenario where there is no shortage of calibration data whatsoever. In particular, we utilize the publicly available MS-COCO dataset \cite{lin2014microsoft}, consisting of images of size $320 \times 240$ depicting various classes to be segmented. We use 20.000 images to calibrate the segmentation model. In our experiments on this dataset, we choose, without loss of generality, the \emph{person} class for our investigations. As is the case for many photographed objects, we expect that `on average' the object of interest will be reasonably centered in the image. We can utilize this knowledge in creating our Kandinsky clusters. In particular, for the genetic and FCC approaches, we will choose them to be concentric sets of circles or ellipses. More precisely, the first cluster, starting from a particular center point, will be a disk, and each subsequent cluster an annulus (or elliptic version thereof). It is also in the formation of such clusters that they have found their name. 

In this extreme high-data calibration regime, the k-means method provides an elucidating insight into the presence of these clusters. In \autoref{fig:kandinsky_cluster_kmeans}, we show the Kandinsky clusters found by the k-means method (art aficionados will recognize the motivation for our method's chosen name in this figure). The presence of these clusters arises from a combination of different prevalences of the persons on the pixels and the difficulty of identifying them. In particular, we expected and found the center of the image to contain most instances of the person class, subsequently decreasing in concentric shapes. If a person is near the boundaries, they are also more likely to be challenging to determine, as they can be out of focus, in the distance, etc.

In \autoref{fig:violin_mscoco_xl}, we show the violin plot of the pixelwise coverage errors for all described methods. In this experiment, and this one only, the baseline pixelwise calibration method slightly outperforms all other methods. This is because, with such an enormous amount of calibration data, we can properly calibrate the individual pixel classifiers. The Kandinsky methods still perform well in this situation, only the image-wise calibration is significantly worse. The moral of this experiment is that provided you have access to tens of thousands of calibration images, pixelwise calibration will perform best, followed closely by all Kandinsky methods.

\subsection{MS-COCO-S}
We now move on to the scenario with little calibration data to spare. This situation is more common, especially in the medical field, due to the cost of annotating large amounts of data. To investigate the low-data scenarios, we now utilize only 100 calibration data points. As one can imagine, the k-means approach will have trouble finding relevant clusters in the images with such few samples. However, our prior knowledge does not change, and we proceed to find the annuli using the GenAnn and FCC approaches, described in \autoref{sec:kandinsky_calibration}. In \autoref{fig:cluster_plot_100}, we show the k-means and the genetic algorithm approach results. It can be seen that the k-means clusters are no longer as clear as in \autoref{fig:kandinsky_cluster_kmeans}. The results of all methods are shown in \autoref{fig:violin_mscoco_s} and \autoref{tab:mean-ces}. In this low-data regime, all Kandinsky methods outperform both pixelwise and imagewise calibration. As expected, the GenAnn and FCC approaches outperform the k-means approach, as they can better incorporate prior knowledge; k-means gets slightly worse due to the noise in the non-conformity curves. 

\begin{figure}
    \centering
    \includegraphics[height=2.7cm]{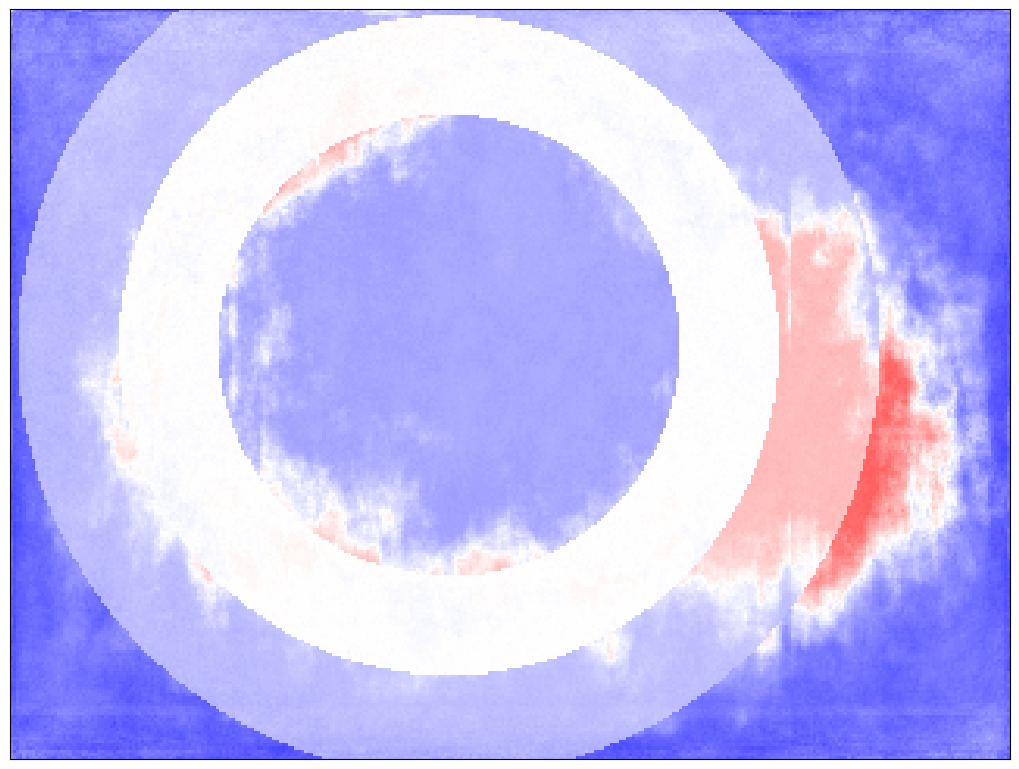} 
    \hfill
    \includegraphics[height=2.7cm]{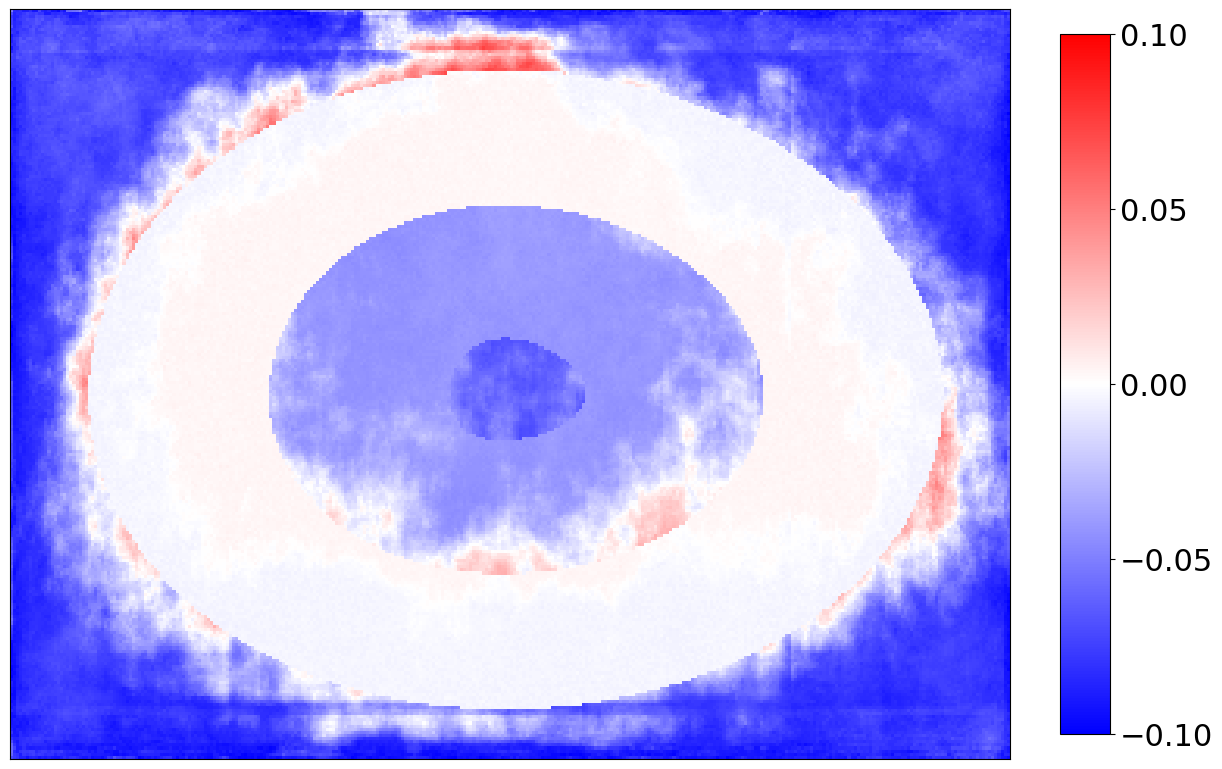} 
    \caption{Subtraction image of coverage errors per pixel for imagewise vs.\ GenAnn calibration (left) and imagewise vs.\ FCC-calibration (right) on the MS-COCO-S dataset. Red indicates lower coverage error for the former, and blue indicates lower coverage error for the latter.}
    \label{fig:ce_subtraction}
\end{figure}

In \autoref{fig:ce_subtraction}, we show the difference in coverage error for the two annuli methods, GenAnn and FCC, versus the errors obtained by imagewise calibration. The Kandinsky methods can group the relevant pixels to get better estimates of non-conformity curves. 


\begin{table*}[ht]
\centering
\begin{sc}
\scalebox{0.99}{
\begin{tabular}{lcccc}
\toprule
\textbf{Method} & \textbf{MS-COCO-XL} & \textbf{MS-COCO-S} & \textbf{Decathlon-L} & \textbf{Decathlon-S} \\ 
\midrule
Pixel           & \textbf{0.055} \small{[0.037, 0.080]} & 0.131 \small{[0.055, 0.319]} & 0.217 \small{[0.041, 0.500]} & 0.249 \small{[0.051, 0.500]} \\ 
Image           & 0.090 \small{[0.038, 0.175]} & 0.091 \small{[0.039, 0.180]} & 0.177 \small{[0.038, 0.410]} & 0.175 \small{[0.036, 0.410]} \\ 
\midrule
K-means         & 0.066 \small{[0.043, 0.114]} & 0.086 \small{[0.039, 0.169]} & \textbf{0.145} \small{[0.038, 0.285]} & 0.168 \small{[0.040, 0.378]} \\ 
Gen-Ann         & 0.067 \small{[0.040, 0.113]} & 0.068 \small{[0.036, 0.134]} & 0.150 \small{[0.041, 0.352]} & \textbf{0.152} \small{[0.037, 0.340]} \\ 
FCC-Ann             & 0.064 \small{[0.037, 0.112]} & \textbf{0.060} \small{[0.032, 0.111]} & 0.152 \small{[0.040, 0.410]} & 0.165 \small{[0.036, 0.410]} \\ 
\bottomrule

\end{tabular}
}
\end{sc}
\caption{Mean coverage errors (with [0.05, 0.95] quantiles) over all pixels in the test subset of four datasets, computed for five distinct calibration methods. The last three methods show our novel methods; all attempt to cluster similar pixels for simultaneous calibration. All Kandinsky clustering-type methods outperform both pixelwise and imagewise calibration in all datasets except MS-COCO-XL, where the calibration set is so large that pixelwise calibration is superior.}
\label{tab:mean-ces}
\end{table*}

\subsection{Decathlon-L}

\begin{figure}
    \centering
    \includegraphics[width=0.8\columnwidth]{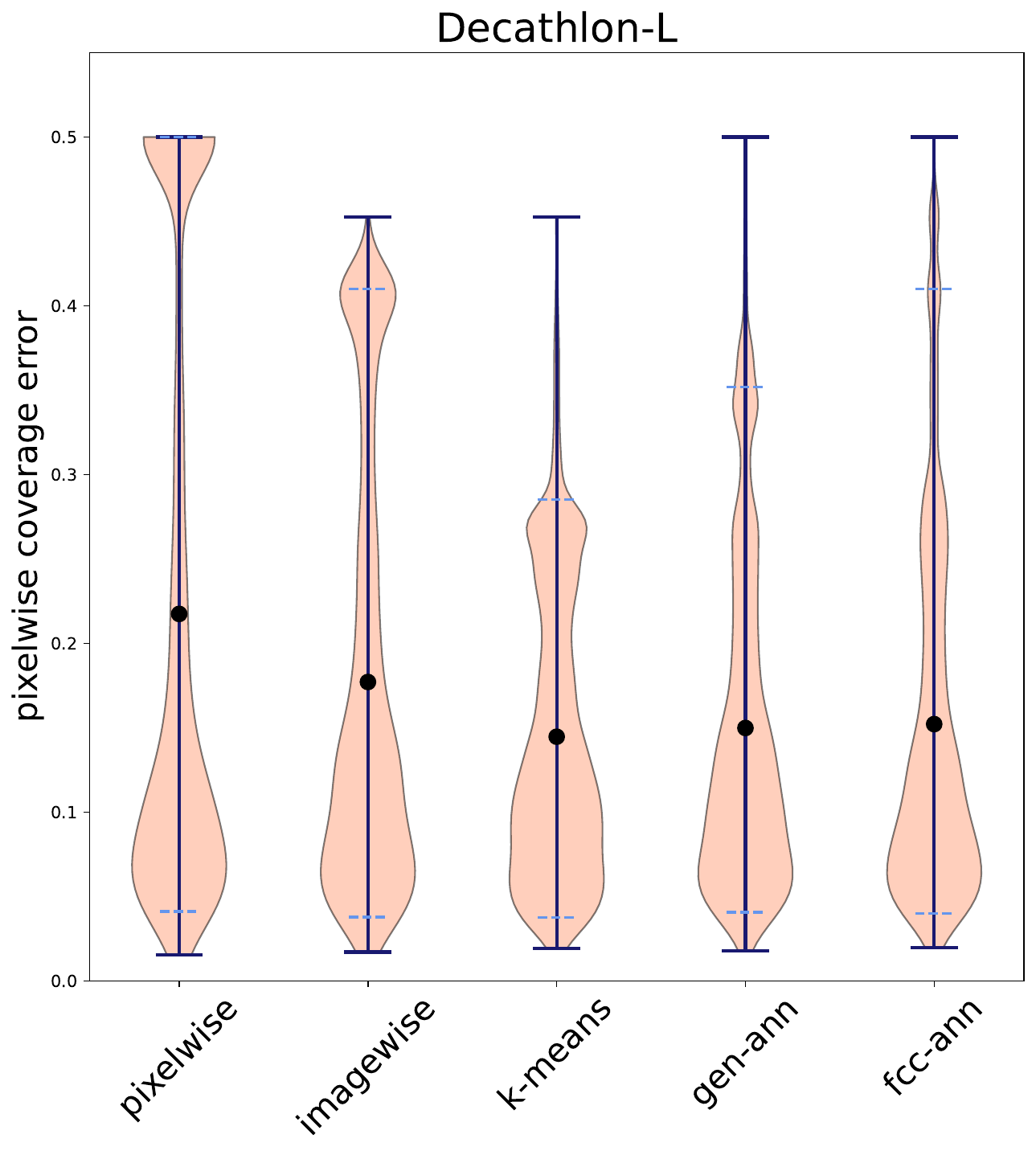}
    \caption{On the large calibration set of 77 patients, all Kandinsky methods have lower coverage errors than the baseline methods. In this case, the K-means clusters can find the most informative clusters of pixels for calibration, closely followed by the annuli methods.}
    \label{fig:violin_md_l}
\end{figure}

We now consider a different segmentation task, falling in the medical domain. In particular, we consider the Medical Decathlon Challenge data \cite{antonelli2022medical}. The object of interest in our case is the Pancreas data, consisting of CT images along with segmentations of the pancreas and (possibly) tumors. All CT scans were sliced, utilizing slices that contained annotations. The data split was made on the patient level. The resulting slices of scans were of size $(384,384)$. For our first experiment, we created a large, for medical standards, calibration dataset consisting of 77 patients. In \autoref{fig:violin_md_l}, we show the violin plot for all methods. As also found in \autoref{tab:mean-ces}, all Kandinsky calibration techniques outperform the pixelwise and imagewise calibration. In this case, the k-means can find the most effective clusters, closely followed by both annuli methods.

\begin{figure}
    \centering
    \includegraphics[width=0.8\columnwidth]{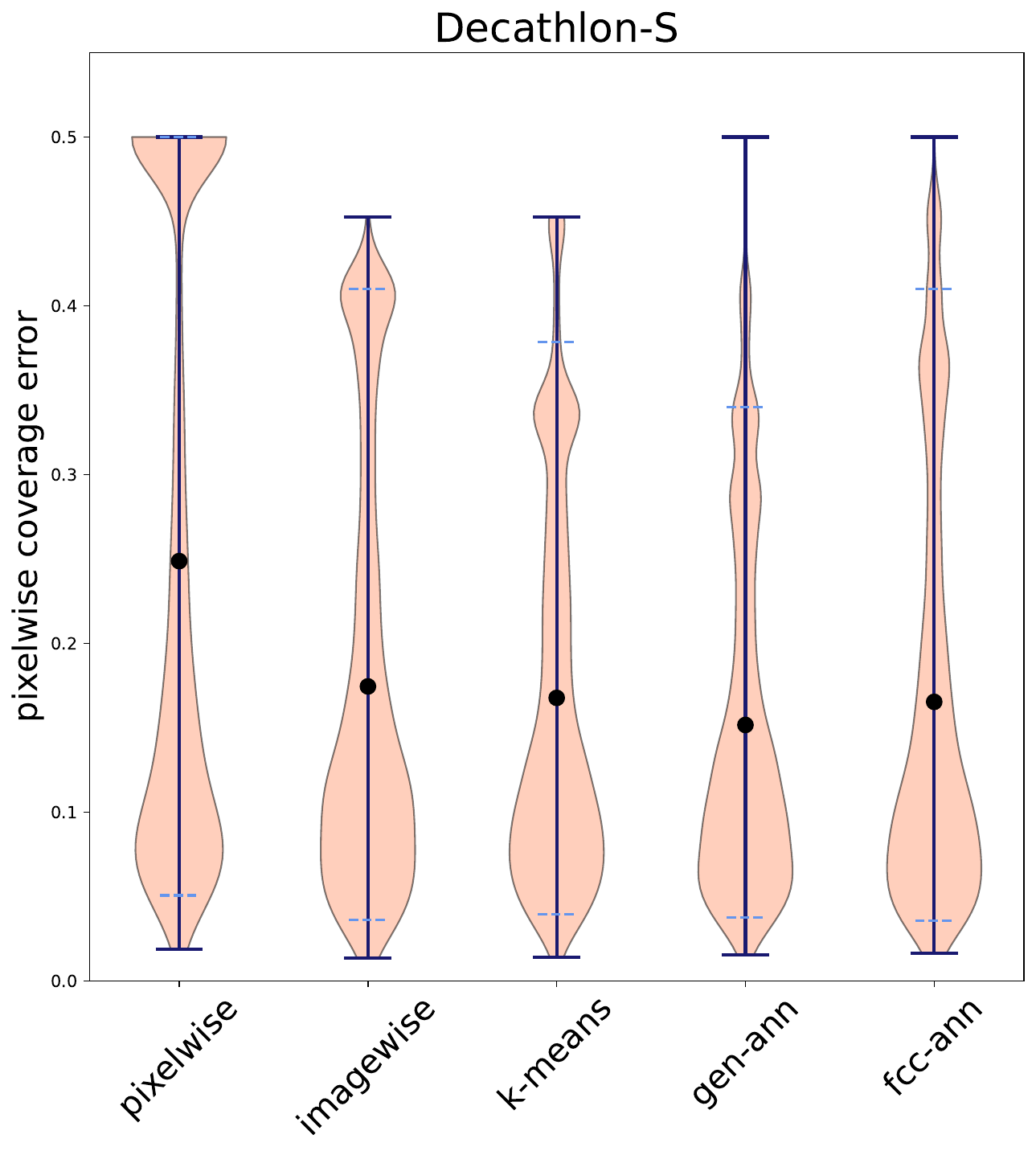}
    \caption{Results for the small decathlon calibration set, using 27 patients. In this case, the GenAnn clusters can most effectively group relevant pixels, closely followed by the k-means and FCC methods. Especially the pixelwise calibration performs poorly in this low-data regime.}
    \label{fig:violin_md_s}
\end{figure}

\subsection{Decathlon-S}
Let us now consider the scenario of little calibration data, which occurs frequently in real-life medical datasets. In particular, we utilize only 27 of the patients for the calibration.

\begin{figure}
    \centering
    \includegraphics[width=0.75\columnwidth]{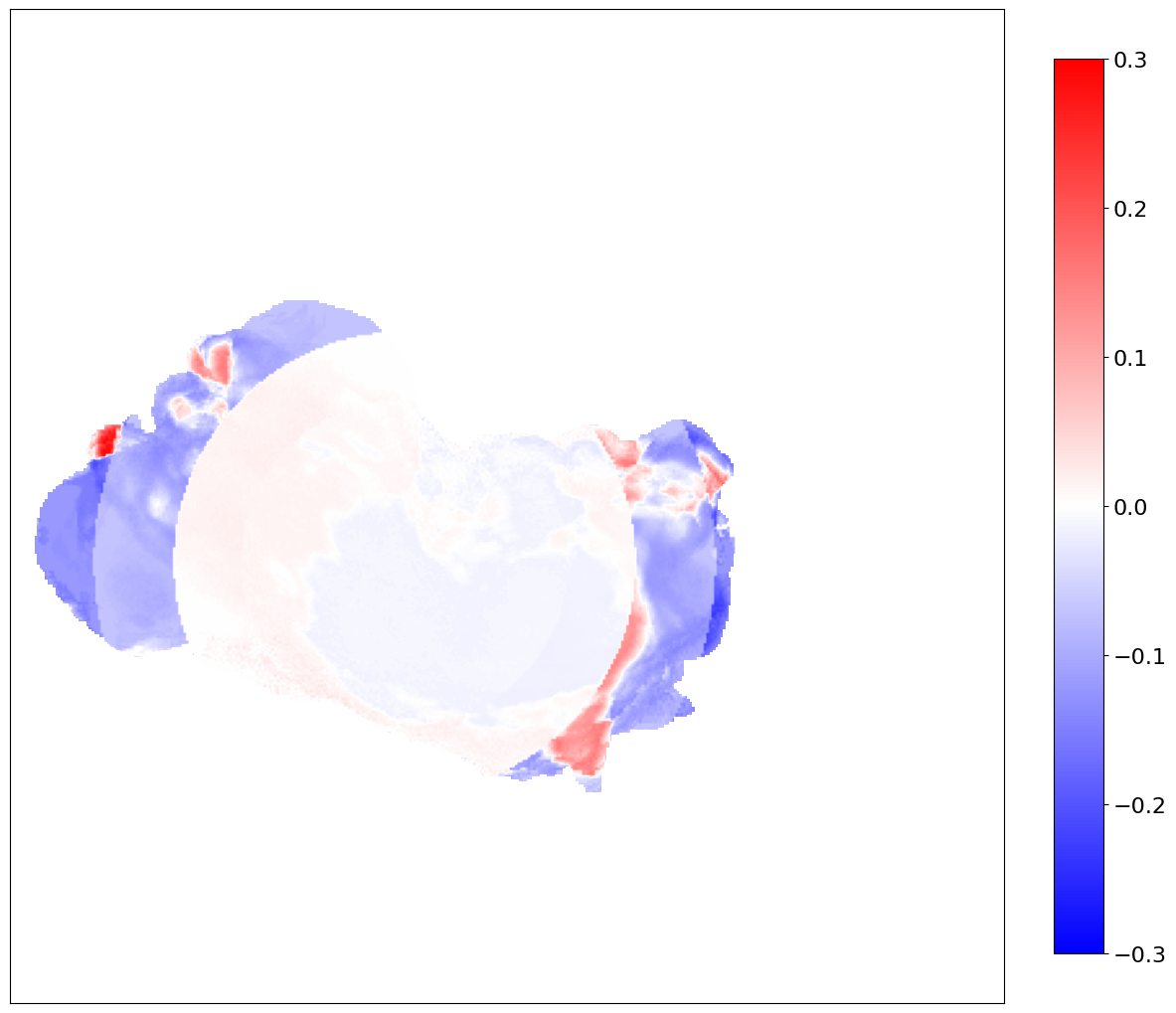}
    \caption{Subtraction image of coverage errors per pixel for imagewise vs. GenAnn calibration on the Decathlon-S dataset. Red indicates lower coverage error for the former, and blue indicates lower coverage error for the latter. Especially near the boundaries, the coverage error decreases for the GenAnn calibration.}
    \label{fig:ce_subtraction_md}
\end{figure}

Due to the increased noise in the calibration procedure, we can expect the k-means clusters to be less meaningful, as they will cluster based on more spurious relations. The methods using prior knowledge, GenAnn and FCC, will retain their strength, as their baked-in knowledge is more robust to the noise in the calibration. The results are shown in \autoref{fig:violin_md_s}. As before, all Kandinsky methods perform better than the pixel and imagewise calibrations. In this case, the GenAnn approach achieves the lowest coverage errors. Furthermore, in \autoref{fig:ce_subtraction_md}, we show the difference in coverage error between the GenAnn and imagewise calibration. Especially near the boundaries, where the pancreas is less often seen, the clusters can find better calibration.

\section{Conclusion}
In this article, we have investigated the problem of efficient calibration for segmentation models. In particular, in practice, there is often little `extra' data available to calibrate models. We have presented progress for solving this problem by presenting the Kandinsky calibration framework. The framework utilizes the well-organized patterns between pixels-classifiers in segmentation networks. We have proposed three methods for computing these clusters in fairly general settings. In all investigated low-data calibration settings, the Kandinsky methods can produce better-calibrated methods, leading to lower coverage errors on unseen data.

{
    \small
    \bibliographystyle{ieeetr}
    \bibliography{bibliography}
}

\clearpage
\appendix
\twocolumn[\section*{Appendix}]
\section{Additional Cluster Plots}
\begin{figure}
	\centering
	{\includegraphics[width=0.9\columnwidth]{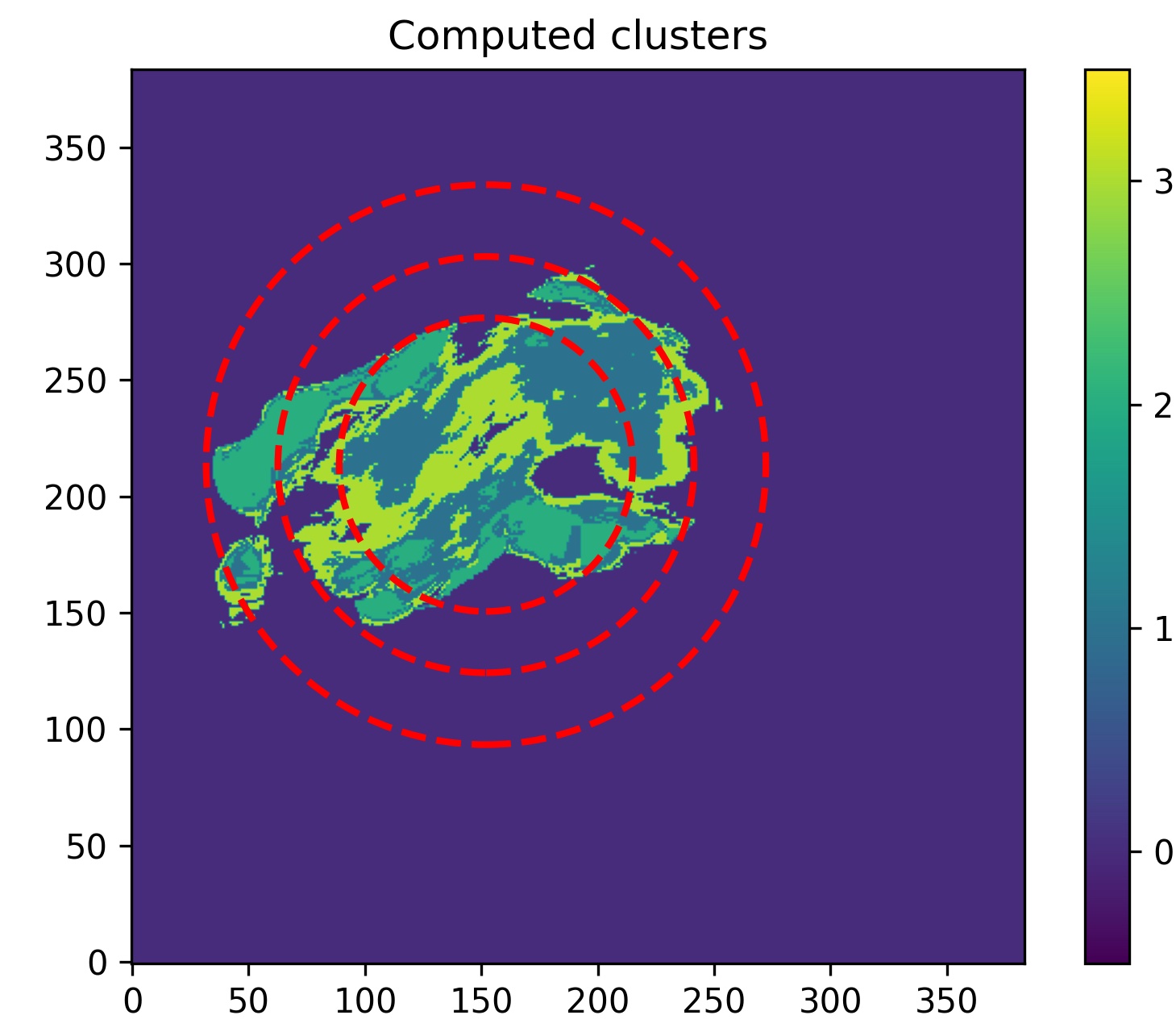}} \\[1ex]
	\caption{The computed k-means and GenAnn annuli clusters for the Medical Decathlon dataset with a small calibration set. 
            \label{fig:cluster_plot_pancreas_low}
		   }
\end{figure}

In \autoref{fig:cluster_plot_pancreas_low} we provide an additional example of k-means clusters and GenAnn clusters, for the Decathlon-S experiment.
\section{Details Genetic Algorithm}\label{appendix:genetic}
Let us start by providing a short insight into the workings of this approach. A candidate solution in a generation $t$, $x^{(t)}_i$, will consist of a finite set of real numbers, the set of $(x,y)$ coordinates for a center, along with a set of radii for annuli clusters. The fitness function we utilize is the evaluation of the ``internal'' distances in each cluster. Each pixel $p$ in a cluster of pixels will have an associated non-conformity curve $n_{p}$, which arises during the calibration procedure. The internal distance $I_C$ for a cluster $C$ of pixels is then defined as
\begin{equation}
   I_C:= \sum_{p,q \in C} d(n_p, n_q) \ ,
\end{equation}
where $d$ is some metric between the two non-conformity curves. We measure the distance between non-conformity scores by comparing the values of both curves at a chosen set of quantiles. Let us denote $D_i^{(t)}$ as the set of clusters of pixels parameterized by a candidate solution, then the fitness function becomes
\begin{equation}
    F\left(x_i^{(t)}\right):= \sum_{C \in D_i^{(t)}} I_C \ .
\end{equation}
We detail the further specifics of the mutation, crossover, and replacement in the supplementary material, \autoref{appendix:genetic}. The main advantage of this method is that provided you can find a parametrization of the clusters you wish to find, it can be utilized.
We provide an overview of the algorithm process in \autoref{algorithm:genetic_algo}.

\begin{algorithm}
  \caption{\label{algorithm:genetic_algo}Differential Evolution Optimization}
  \begin{algorithmic}[1]
    \State Initialize population with random candidate solutions $P^{(0)} = \{x_1^{(0)}, x_2^{(0)}, ..., x_N^{(0)}\}$
    \State Evaluate the fitness $f_i^{(0)} = F(x_i^{(0)})$ for each individual
    \For{each generation $t = 1, 2, ..., T$}
      \State Apply mutation $P_{\text{mut}}^{(t)} = M(P^{(t-1)})$
    \State Create offspring via crossover $P_{\text{cross}}^{(t)} = C(P_{\text{mut}}^{(t)})$
      \State Replace population $P^{(t+1)} = R(P^{(t-1)}, P_{\text{cross}}^{(t)})$
      \State Re-evaluate the fitness for the new population
      \State Check for termination condition 
    \EndFor
    \State Output the best solution $x^* = \argmin_{x_i^{(T^*)}} f_i^{(T^*)}$ with $T^*$ the last generation
  \end{algorithmic}
\end{algorithm}

For the genetic algorithm approach of finding annuli in the MS-COCO case, we have used the range $[-20,20]$ as bounds for the center $x$ and $y$ coordinates, and $[0,150]$ for the two innermost radii and $[0,300]$ for the outermost (as it is allowed to lie outside of the image if necessary). In the case of the medical decathlon dataset, the radii, the bounds of the center coordinates were $[-60,60]$ and the inner radii up to $200$ and outermost $400$. These choices were made based on the image sizes. We did not find that it was necessary to add any `repulsion' term in the objective function between different radii, as it is generally a more favorable solution to have separate annuli for lower objective function values.\\
For the mutation and crossover, we start by selecting three random individuals from the population, $a$, $b$, and $C$. The mutation will create a new, mutated individual by
\begin{equation}
    x_{\text{mut}} = a + C_M(b-c) \ ,
\end{equation}
where $C_M$ denotes the mutation factor, chosen to be $0.8$ in our experiments. The operations in the mutation are pointwise on the individuals. Subsequently, a crossover will be performed between the candidate solution and the mutated one. Suppose each candidate solution is of length $n$, i.e., it has $n$ parameters specifying clusters. We first define a crossover probability $p_C=0.7$ and then uniformly sample $n$ random numbers that lie in $[0,1)$. Whenever the random number exceeds $p_C$, we accept the mutated parameter. After the mutation and crossover, we end up with a new candidate solution $\Tilde{x}$, which replaces the original candidate solution $x$ only if the fitness is lower.

We note that the computation of the `internal' distance, i.e., the distance between the quantiles of the non-conformity curves, is very amenable to large matrix operations. Therefore, we have implemented it using \textsc{PyTorch}, allowing for a batched, GPU-compatible computation. All our computations were performed on an Nvidia A6000 GPU but can be on a CPU as well.

\section{Details Fourier Concentric Clustering}\label{appendix:fcc}
In this section we explain how to construct a concentric decomposition of $\mathcal{R}_{n} := [0, n_{1} - 1] \times [0, n_{2} -1]$ on which 
the variances of a user-prescribed quantity $\J: \mathcal{R}_{n} \rightarrow \RR^{p}$ are minimized. Here 
$n := (n_{1}, n_{2}) \in \NN^{2}$ corresponds to the measurements of the domain of the underlying image of interest. 
Specifically, we explain how to construct a sequence of simply connected nested subsets 
$
	V_{0} \subset \ldots \subset V_{m-1} \subset \mathcal{R}_{n}
$
such that $\sum_{l=0}^{m} \E_{A_{l}} \left( \left \Vert \J - \bm{\mu}_{l} \right \Vert^{2} \right)$ is minimal, where 
\begin{align*}
	A_{l} := 
	\begin{cases}
		V_{0}, & l = 0, \\
		V_{l} \setminus V_{l-1}, & 1 \leq l \leq m-1, \\
		\mathcal{R}_{n} \setminus V_{m-1}, & l = m.
	\end{cases}
\end{align*}
Here $m \in \NN$ is the number of sets in the filtration $(V_{l})_{l=0}^{m-1}$, 
$\Vert \cdot \Vert$ denotes the standard Euclidean norm on $\RR^{p}$ and 
$\bm{\mu}_{l} = \E_{A_{l}}( \J)\in \RR^{p}$ is the mean of $\J$ on $A_{l}$. We require that each set $V_{l}$ has a smooth 
boundary and refer to $A_{l}$ as a domain. 

\subsection{Explicit representation of the filtration}
We start by constructing an explicit representation of each $A_{l}$ amenable to numerical computations. 
First, define a new (Cartesian) coordinate system in which the midpoint 
$\bm{p} = (p_{1}, p_{2})$ of $\mathcal{R}_{n}$ corresponds to the origin. Next, parameterize $A_{0}$ by assuming its boundary
$\partial A_{0}$ is a polar curve, i.e., in polar coordinates we have 
\begin{align*}
	A_{0} = 
	\left \{ (r, \theta): 0 \leq r \leq r_{0}(\theta), \ \theta \in [0, 2 \pi] \right \}, 
\end{align*}
where $r_{0} : [0, 2\pi] \rightarrow (0, \infty)$ is a continuously differentiable $2\pi$-periodic map. 
Similarly, we assume $r_{l-1}, r_{l}: [0, 2 \pi] \rightarrow (0, \infty)$ are polar curves parameterizing the boundaries of 
$V_{l-1}$ and $V_{l}$, respectively, and hence
\begin{align*}
	A_{l} = \left \{ (r, \theta) \right. &\left.: r_{l-1}(\theta) \leq r \leq r_{l}(\theta), \ \theta \in [0, 2 \pi] \right \}, \\
	&\quad 1 \leq l \leq m -1.
\end{align*}
For the final domain $A_{m}$, the ``outer'' boundary $\partial \mathcal{R}_{n}$ is fixed and no
explicit parameterization is required:
\begin{align*}
	A_{m} =  &\left \{ (r, \theta): r \geq r_{m-1}(\theta),  \
				  0 \leq p_{1} + r \cos \theta \leq n_{1} -1,  \right. \\[2ex] & \left. 
				  \quad 0 \leq p_{2} + r \sin \theta \leq n_{2} -1, \ 
				  \theta \in [0, 2 \pi]
				  \right \}. 
\end{align*}

The domains $(A_{l})_{l=0}^{m}$ are fully determined by the polar curves
$(r_{l})_{l=0}^{m-1}$. Since these mappings are smooth and $2\pi$-periodic, 
they admit uniformly convergent Fourier series. For this reason, we have chosen 
to represent the polar curves using finite Fourier expansions:
\begin{align*}
	r_{l}(\theta) := \sum_{\vert k \vert \leq K - 1} a_{lk} e^{i k \theta}, \quad 0 \leq l \leq m -1, 
\end{align*}
where $(a_{lk})_{k = 1 - K}^{K-1} \in \CC^{2 K - 1}$ are the Fourier coefficients of $r_{l}$. 
Here $K \in \NN$ is a user-prescribed hyperparameter which we shall refer to as
the order of the Fourier series. We only store the Fourier coefficients 
$a_{l} := (a_{lk})_{k=0}^{K-1} \in \CC^{K}$, since $r_{l}$ is real-valued. 

\begin{remark}
	In practice, we actually define
	\begin{align*}
		r_{l}(\theta) := \left(  \sum_{\vert k \vert \leq K - 1} a_{lk} e^{i k \theta} \right)^{2}, 
        \quad 0 \leq l \leq m -1,
	\end{align*}
	to ensure that the resulting radii are positive. For the sake of presentation, however, we shall
	ignore this minor detail in the following sections. 
\end{remark}

\subsection{Numerical evaluation of integrals}
To compute the variances $ \E_{A_{l}} \left( \left \Vert \J - \bm{\mu}_{l} \right \Vert^{2} \right)$, as (differentiable) functions of 
$\bm{a} := \left( a_{0}, \ldots, a_{m-1} \right) \in \CC^{mK}$, we need to numerically 
evaluate integrals over the domains $A_{l}$. We use a combination of Legendre and Fourier 
quadrature to accomplish this. In this section we explain in detail how to numerically 
integrate an arbitrary continuously differentiable map $f: \mathcal{R}_{n} \rightarrow \RR$ over each domain. 

For the innermost domain $A_{0}$, we have
\begin{align*}
	\int_{A_{0}} &f(x) \dx = \\
 &\int_{0}^{2 \pi} \int_{0}^{r_{0}(\theta)} f \left( p_{1} + r \cos \theta, p_{2} + r \sin \theta \right) r \dr \dtheta. 
\end{align*}
Notice that the inner integral is a $2 \pi$-periodic function of $\theta$ and continuously differentiable. 
Therefore, it admits a unique Fourier expansion. In particular, if $\left( c_{0k} \right)_{k \in \ZZ}$ are the corresponding 
Fourier coefficients, then 
\begin{align}
	\label{eq:approx_int}
	\int_{A_{0}} f(x) \dx = 2 \pi c_{00}. 
\end{align}
Consequently, to approximate the desired integral, it suffices to approximate the zero$^{\text{th}}$ Fourier coefficient of
\begin{align}
	\label{eq:approx_line_int}
	\theta \stackrel{I_{0}}{\mapsto} \int_{0}^{r_{0}(\theta)} f \left( p_{1} + r \cos \theta, p_{2} + r \sin \theta \right) r \dr.
\end{align}
For this purpose, we first sample \eqref{eq:approx_line_int} on an equidistributed grid of 
$[0, 2 \pi]$ of size $K_{I} \in \NN$. To approximate the integrals in \eqref{eq:approx_line_int} at
fixed angles, we use Gaussian quadrature of order $K_{G} \in \NN$. Next, we use the (inverse) Fast Fourier Transform (FFT) 
to approximate $(c_{0l})_{l=0}^{K_{I} -1}$, and in turn the desired integral using \eqref{eq:approx_int}. 

The integrals for $1 \leq l \leq m-1$ are approximated in a similar fashion. More precisely, for $1 \leq l \leq m-1$, we have 
\begin{align*}
	\int_{A_{l}} &f(x) \dx =\\
 &\int_{0}^{2 \pi} \int_{r_{l-1}(\theta)}^{r_{l}(\theta)} f \left( p_{1} + r \cos \theta, p_{2} + r \sin \theta \right) r \dr \dtheta. 
\end{align*}
As before, we first note that
\begin{align*}
	\label{eq:approx_line_int_annuli}
	\theta \stackrel{I_{l}}{\mapsto} \int_{0}^{2 \pi} \int_{r_{l-1}(\theta)}^{r_{l}(\theta)} f \left( p_{1} + r \cos \theta, p_{2} + r \sin \theta \right) r \dr \dtheta
\end{align*}
is a continuously differentiable $2 \pi$-periodic function. Hence it admits a unique Fourier expansion. Therefore, if
$\left( c_{lk} \right)_{k \in \ZZ}$ are the associated Fourier coefficients, then 
\begin{align*}
	\int_{A_{l}} f(x) \dx = 2 \pi c_{l0}, \quad 1 \leq l \leq m -1,
\end{align*}
as before, and we approximate the right-hand side in exactly the same way. 

Finally, we approximate the integral over $A_{m}$ using a combination of two-dimensional quadrature and the
computations above. More precisely, first observe that 
\begin{align*}
	\int_{A_{m}} f(x) \dx = \int_{\mathcal{R}_{n}} f(x) \dx - \int_{V_{m-1}} f(x) \dx. 
\end{align*}
We approximate the first integral using $2$-dimensional Gaussian quadrature with order $(K_{G}, K_{G}) \in \NN^{2}$. 
The second integral is approximated using the strategy explained in the previous paragraph. 

\begin{remark}
	One could use different orders $K$, $K_{I}$, $K_{G}$ (and $(K_{G}, K_{G})$ for the final domain)
	on each domain. For simplicity, to reduce the number of hyperparameters, we have kept them the
	same across all domains. 
\end{remark}

\subsection{Numerical evaluation of variances}
Next, we explain how to compute $\E_{A_{l}} \left( \left \Vert \J - \bm{\mu}_{l} \right \Vert^{2} \right)$ given Fourier coefficients $\bm{a}$. To this end, first observe
that the area of $A_{l}$ is given by 
\begin{align*}
\scalebox{0.82}{
$\begin{aligned}
\lambda(A_{l}) = 
\begin{cases}
    \dfrac{1}{2} \displaystyle \int_{0}^{2 \pi} r_{0}(\theta)^{2} \, d\theta, & l = 0, \\[2ex]
    \dfrac{1}{2} \displaystyle \int_{0}^{2 \pi} \left( r_{l}(\theta)^{2} - r_{l-1}(\theta)^{2} \right) \, d\theta, & 1 \leq l \leq m -1, \\[2ex]
    (n_{2} - 1) (n_{1} - 1)-  \dfrac{1}{2}\displaystyle \int_{0}^{2 \pi} r_{m-1}(\theta)^{2} \, d\theta, & l = m,
\end{cases}
\end{aligned}$
}
\end{align*}
where $\lambda$ is the Lebesgue measure on $\mathcal{R}_{n}$. Using the Fourier representations of the polar curves
$r_{l}$, we see that 
\begin{align*}
\scalebox{0.82}{
$\begin{aligned}
	\lambda(A_{l}) = 
	\begin{cases}
		\pi \left( a_{0} \ast a_{0} \right)_{0}, & l = 0, \\[2ex]
		\pi  \left( 
            \left( a_{l} \ast a_{l} \right)_{0} - \left( a_{l-1} \ast a_{l-1} \right)_{0} 
        \right), & 1 \leq l \leq m -1, \\[2ex]
		(n_{2} - 1) (n_{1} - 1) - \pi \left( a_{m-1} \ast a_{m-1} \right)_{0}, & l = m,
	\end{cases}
 \end{aligned}$
}
\end{align*}
where $\ast$ denotes the two-sided discrete convolution. We use these computations to turn each $A_{l}$ into a 
probability space $\left( A_{l}, \mathcal{B} \left( A_{l} \right), \PP_{l} \right)$, where $ \mathcal{B} \left( A_{l} \right)$ 
is the Borel-sigma algebra on $A_{l}$, and  $\PP_{l} = \frac{\lambda \vert_{A_{l}}}{\lambda(A_{l})}$. 

We may now readily use the approximation techniques described in the previous section to approximate the variance 
of $\J$ on $A_{l}$: 
\begin{align*}
	\E_{A_{l}} \left( \left \Vert \J - \bm{\mu}_{l} \right \Vert^{2} \right) &= 
	\frac{1}{ \lambda( A_{l} ) } \int_{A_{l}} \left \Vert \J(x) - \bm{\mu}_{l} \right \Vert^{2}  \dx, \\
	 \bm{\mu}_{l}  &=  \frac{1}{ \lambda( A_{l} ) } \int_{A_{l}} \J(x) \dx
\end{align*}
for $0 \leq l \leq m$. 

\subsection{Setting up a minimization problem}
Finally, we set up a minimization problem to find a concentric decomposition of $\mathcal{R}_{n}$. 
We choose the initial filtration $(V_{l})_{l=0}^{m-1}$ as a ``noisy'' set of concentric nested
circles. This corresponds to the following initialization of the Fourier coefficients of $(r_{l})_{l=0}^{m-1}$:
\begin{align*}
	a_{l} = (R_{l}, 0, \ldots, 0) + \varepsilon, \quad 0 \leq l \leq m-1,
\end{align*}
where $\varepsilon \sim N(\bm{0}_{K}, \sigma \bm{I}_{K \times K})$, $\sigma >0$,
$R_{l} = \frac{l}{m} R_{\max}$, and $R_{\max} >0$ is a hyperparameter chosen so that the circle with radius $R_{\max}$
and midpoint $\bm{p}$ is contained in $\mathcal{R}_{n}$. With this initialization, we use the BFGS-algorithm to solve 
\begin{align*}
	\min_{\bm{a} \in \CC^{mK}} \left( \sum_{l=0}^{m} \E_{A_{l}} \left( \left \Vert \J - \bm{\mu}_{l} \right \Vert^{2} \right) + w \sum_{l=1}^{m} \Lambda(\bm{a})_{l} \right), 
\end{align*}
where $w >0$ is a fixed weight, and 
\begin{align*}
	\Lambda( \bm{a} )_{l} := 
	\begin{cases}
		\displaystyle \int_{0}^{2 \pi} \frac{1}{ r_{l}(\theta) - r_{l-1}(\theta) } \dtheta, & 1 \leq l \leq m-1, \\[2ex]
		\displaystyle \int_{0}^{2 \pi} \frac{1}{ R_{\max} - r_{m -1}(\theta) } \dtheta, & l = m,
	\end{cases}
\end{align*}
is a penalty term that enforces the sets $V_{l}$ to remain nested. That is, it enforces that the boundaries do not touch each other. 
The integrals are approximated, as before, by computing the zero$^{\text{th}}$ Fourier coefficients of the integrands, which in turn
are obtained by sampling them on an equispaced grid of $[0, 2 \pi]$ of size $K_{\Lambda} \in \NN$ and using the (inverse) FFT. 

\begin{remark}
	The penalty term for $l=m$ could be improved substantially by explicitly parameterizing $\partial \mathcal{R}_{n}$. 
	In particular, the current term yields a suboptimal filtration, since it confines the final polar curve $r_{m-1}$ to a circle
	incscribed in $\mathcal{R}_{n}$. 
\end{remark}

\end{document}

%% file: preamble.tex
\usepackage{graphicx}
\usepackage{booktabs} 

\usepackage{url}
\usepackage{times}
\usepackage{algorithm}
\usepackage{algpseudocode}
\usepackage{amsmath}
\usepackage{amssymb}
\usepackage{bm}
\usepackage{mathtools}
\usepackage{amsthm}
\usepackage{esint}

\theoremstyle{plain}
\newtheorem{theorem}{Theorem}[section]

\theoremstyle{definition}

\theoremstyle{remark}
\newtheorem{remark}[theorem]{Remark}

\usepackage{bm}
\usepackage{enumerate}
\newcommand{\RR}{\mathbb R}
\newcommand{\ZZ}{\mathbb Z}
\newcommand{\CC}{\mathbb C}
\newcommand{\NN}{\mathbb N}

\newcommand{\PP}{\mathbb{P}}
\newcommand{\E}{\mathbb{E}}

\newcommand{\J}{\mathcal{J}}

\newcommand{\dx}{\ \mbox{d}x}

\DeclareMathOperator*{\argmin}{arg\,min}

\DeclareMathOperator{\dr}{\text{d}r}
\DeclareMathOperator{\dtheta}{\text{d}\theta}

\usepackage[dvipsnames]{xcolor}

%% file: main.bbl
\begin{thebibliography}{10}

\bibitem{angelopoulos2022conformal}
A.~N. Angelopoulos, S.~Bates, A.~Fisch, L.~Lei, and T.~Schuster, ``Conformal {{Risk Control}},'' Aug. 2022.

\bibitem{angelopoulos2022learn}
A.~N. Angelopoulos, S.~Bates, E.~J. Cand{\`e}s, M.~I. Jordan, and L.~Lei, ``Learn then {{Test}}: {{Calibrating Predictive Algorithms}} to {{Achieve Risk Control}},'' Sept. 2022.

\bibitem{angelopoulos2023predictionpowered}
A.~N. Angelopoulos, S.~Bates, C.~Fannjiang, M.~I. Jordan, and T.~Zrnic, ``Prediction-{{Powered Inference}},'' Feb. 2023.

\bibitem{kutiel2023conformal}
G.~Kutiel, R.~Cohen, M.~Elad, D.~Freedman, and E.~Rivlin, ``Conformal {{Prediction Masks}}: {{Visualizing Uncertainty}} in {{Medical Imaging}},'' in {\em {{ICLR}} 2023 {{Workshop}} on {{Trustworthy Machine Learning}} for {{Healthcare}}}, Apr. 2023.

\bibitem{romano2020classification}
Y.~Romano, M.~Sesia, and E.~Candes, ``Classification with {{Valid}} and {{Adaptive Coverage}},'' in {\em Advances in {{Neural Information Processing Systems}}}, vol.~33, pp.~3581--3591, {Curran Associates, Inc.}, 2020.

\bibitem{angelopoulos2022imagetoimage}
A.~N. Angelopoulos, A.~P. Kohli, S.~Bates, M.~I. Jordan, J.~Malik, T.~Alshaabi, S.~Upadhyayula, and Y.~Romano, ``Image-to-{{Image Regression}} with {{Distribution-Free Uncertainty Quantification}} and {{Applications}} in {{Imaging}},'' Feb. 2022.

\bibitem{saunders1999transduction}
C.~Saunders, A.~Gammerman, and V.~Vovk, ``Transduction with {{Confidence}} and {{Credibility}},'' in {\em Sixteenth {{International Joint Conference}} on {{Artificial Intelligence}} ({{IJCAI}} '99) (01/01/99)}, pp.~722--726, 1999.

\bibitem{papadopoulos2002inductive}
H.~Papadopoulos, K.~Proedrou, V.~Vovk, and A.~Gammerman, ``Inductive confidence machines for regression,'' in {\em Machine Learning: ECML 2002: 13th European Conference on Machine Learning Helsinki, Finland, August 19--23, 2002 Proceedings 13}, pp.~345--356, Springer, 2002.

\bibitem{papadopoulos2007conformal}
H.~Papadopoulos, V.~Vovk, and A.~Gammerman, ``Conformal {{Prediction}} with {{Neural Networks}},'' in {\em 19th {{IEEE International Conference}} on {{Tools}} with {{Artificial Intelligence}}({{ICTAI}} 2007)}, vol.~2, pp.~388--395, Oct. 2007.

\bibitem{papadopoulos2008inductive}
H.~Papadopoulos, {\em Inductive {{Conformal Prediction}}: {{Theory}} and {{Application}} to {{Neural Networks}}}.
\newblock {IntechOpen}, Aug. 2008.

\bibitem{vovk2012conditional}
V.~Vovk, ``Conditional {{Validity}} of {{Inductive Conformal Predictors}},'' in {\em Proceedings of the {{Asian Conference}} on {{Machine Learning}}}, pp.~475--490, {PMLR}, Nov. 2012.

\bibitem{guo2017calibration}
C.~Guo, G.~Pleiss, Y.~Sun, and K.~Q. Weinberger, ``On {{Calibration}} of {{Modern Neural Networks}},'' Aug. 2017.

\bibitem{minderer2021revisiting}
M.~Minderer, J.~Djolonga, R.~Romijnders, F.~Hubis, X.~Zhai, N.~Houlsby, D.~Tran, and M.~Lucic, ``Revisiting the {{Calibration}} of {{Modern Neural Networks}},'' Oct. 2021.

\bibitem{carrell2022calibration}
A.~M. Carrell, N.~Mallinar, J.~Lucas, and P.~Nakkiran, ``The {{Calibration Generalization Gap}},'' Oct. 2022.

\bibitem{conde2023theoretical}
P.~Conde, R.~L. Lopes, and C.~Premebida, ``A {{Theoretical}} and {{Practical Framework}} for {{Evaluating Uncertainty Calibration}} in {{Object Detection}},'' Sept. 2023.

\bibitem{kuppers2022confidence}
F.~K{\"u}ppers, A.~Haselhoff, J.~Kronenberger, and J.~Schneider, ``Confidence {{Calibration}} for {{Object Detection}} and {{Segmentation}},'' in {\em Deep {{Neural Networks}} and {{Data}} for {{Automated Driving}}: {{Robustness}}, {{Uncertainty Quantification}}, and {{Insights Towards Safety}}} (T.~Fingscheidt, H.~Gottschalk, and S.~Houben, eds.), pp.~225--250, {Cham}: {Springer International Publishing}, 2022.

\bibitem{blasiok2022unifying}
J.~B{\l}asiok, P.~Gopalan, L.~Hu, and P.~Nakkiran, ``A {{Unifying Theory}} of {{Distance}} from {{Calibration}},'' Nov. 2022.

\bibitem{vovk2005algorithmic}
V.~Vovk, A.~Gammerman, and G.~Shafer, {\em Algorithmic learning in a random world}, vol.~29.
\newblock Springer, 2005.

\bibitem{lei2014distribution}
J.~Lei and L.~Wasserman, ``Distribution-free prediction bands for non-parametric regression,'' {\em Journal of the Royal Statistical Society Series B: Statistical Methodology}, vol.~76, no.~1, pp.~71--96, 2014.

\bibitem{angelopoulos2021gentle}
A.~N. Angelopoulos and S.~Bates, ``A gentle introduction to conformal prediction and distribution-free uncertainty quantification,'' {\em arXiv preprint arXiv:2107.07511}, 2021.

\bibitem{wieslander2021deep}
H.~Wieslander, P.~J. Harrison, G.~Skogberg, S.~Jackson, M.~Frid{\'e}n, J.~Karlsson, O.~Spjuth, and C.~W{\"a}hlby, ``Deep {{Learning With Conformal Prediction}} for {{Hierarchical Analysis}} of {{Large-Scale Whole-Slide Tissue Images}},'' {\em IEEE Journal of Biomedical and Health Informatics}, vol.~25, pp.~371--380, Feb. 2021.

\bibitem{blasiok2023smooth}
J.~B{\l}asiok and P.~Nakkiran, ``Smooth {{ECE}}: {{Principled Reliability Diagrams}} via {{Kernel Smoothing}},'' Sept. 2023.

\bibitem{holland1992adaptation}
J.~H. Holland, {\em Adaptation in natural and artificial systems: an introductory analysis with applications to biology, control, and artificial intelligence}.
\newblock MIT press, 1992.

\bibitem{goldberg1989cenetic}
D.~E. Goldberg, ``Cenetic algorithms in search,'' {\em Optimization, Machine Learning}, 1989.

\bibitem{storn1997differential}
R.~Storn and K.~Price, ``Differential evolution--a simple and efficient heuristic for global optimization over continuous spaces,'' {\em Journal of global optimization}, vol.~11, pp.~341--359, 1997.

\bibitem{broyden1970convergence}
C.~G. Broyden, ``The convergence of a class of double-rank minimization algorithms 1. general considerations,'' {\em IMA Journal of Applied Mathematics}, vol.~6, no.~1, pp.~76--90, 1970.

\bibitem{shanno1970conditioning}
D.~F. Shanno, ``Conditioning of quasi-newton methods for function minimization,'' {\em Mathematics of computation}, vol.~24, no.~111, pp.~647--656, 1970.

\bibitem{fletcher1970new}
R.~Fletcher, ``A new approach to variable metric algorithms,'' {\em The computer journal}, vol.~13, no.~3, pp.~317--322, 1970.

\bibitem{goldfarb1970family}
D.~Goldfarb, ``A family of variable-metric methods derived by variational means,'' {\em Mathematics of computation}, vol.~24, no.~109, pp.~23--26, 1970.

\bibitem{lin2014microsoft}
T.-Y. Lin, M.~Maire, S.~Belongie, J.~Hays, P.~Perona, D.~Ramanan, P.~Doll{\'a}r, and C.~L. Zitnick, ``Microsoft coco: Common objects in context,'' in {\em Computer Vision--ECCV 2014: 13th European Conference, Zurich, Switzerland, September 6-12, 2014, Proceedings, Part V 13}, pp.~740--755, Springer, 2014.

\bibitem{antonelli2022medical}
M.~Antonelli, A.~Reinke, S.~Bakas, K.~Farahani, A.~Kopp-Schneider, B.~A. Landman, G.~Litjens, B.~Menze, O.~Ronneberger, R.~M. Summers, {\em et~al.}, ``The medical segmentation decathlon,'' {\em Nature communications}, vol.~13, no.~1, pp.~1--13, 2022.

\end{thebibliography}
